\documentclass[10pt,twocolumn,letterpaper]{article}

\usepackage{iccv}              

\usepackage{microtype} 
\usepackage{xspace}
\usepackage{xcolor}

\newcommand{\sysname}{CINEMA\xspace}

\usepackage{amsmath}

\newcommand{\loss}{\mathcal{L}}
\newcommand{\losscos}{\loss_\mathrm{cos}}
\newcommand{\lossmse}{\loss_\mathrm{mse}}
\newcommand{\videoclip}{V}
\newcommand{\image}{I}
\newcommand{\referenceimage}{\image}
\newcommand{\features}{\mathbf{F}}

\newcommand{\tfivefeatures}{\features_\mathrm{T5}}
\newcommand{\mllmfeatures}{\features_\mathrm{MLLM}}

\newcommand{\visualfeatures}{\features_{\mathrm{visual}}}
\newcommand{\inputfeatures}{\features_{\mathrm{input}}}
\newcommand{\noisetokens}{\features_{\mathrm{noise}}^t}
\newcommand{\unifiedfeatures}{\features_{\mathrm{unified}}}
\newcommand{\baralphat}{\bar{\alpha}_t}

\newcommand{\captions}{C}

\definecolor{iccvblue}{rgb}{0.21,0.49,0.74}

\usepackage[pagebackref,breaklinks,colorlinks,allcolors=iccvblue]{hyperref}

\def\paperID{1152} 
\def\confName{ICCV}
\def\confYear{2025}

\title{CINEMA: Coherent Multi-Subject Video Generation via \\ MLLM-Based Guidance}

\author{Yufan Deng\textsuperscript{1,2} \hspace{0.18in}
Xun Guo\textsuperscript{1} \hspace{0.18in}
Yizhi Wang\textsuperscript{1} \hspace{0.18in}
Jacob Zhiyuan Fang\textsuperscript{1} \hspace{0.18in}
Angtian Wang\textsuperscript{1}
\\
Shenghai Yuan\textsuperscript{2} \hspace{0.2in}
Yiding Yang\textsuperscript{1} \hspace{0.2in}
Bo Liu\textsuperscript{1} \hspace{0.2in}
Haibin Huang\textsuperscript{1} \hspace{0.2in}
Chongyang Ma\textsuperscript{1}
\vspace{4pt}
\\
\textsuperscript{1}{ByteDance Intelligent Creation}\hspace{0.3in} \textsuperscript{2}{Peking Univeristy}
}

\begin{document}

\twocolumn[{%
\renewcommand\twocolumn[1][]{#1}%
\maketitle
\begin{center}
    \centering
    \captionsetup{type=figure}
    \includegraphics[width=0.98\linewidth]{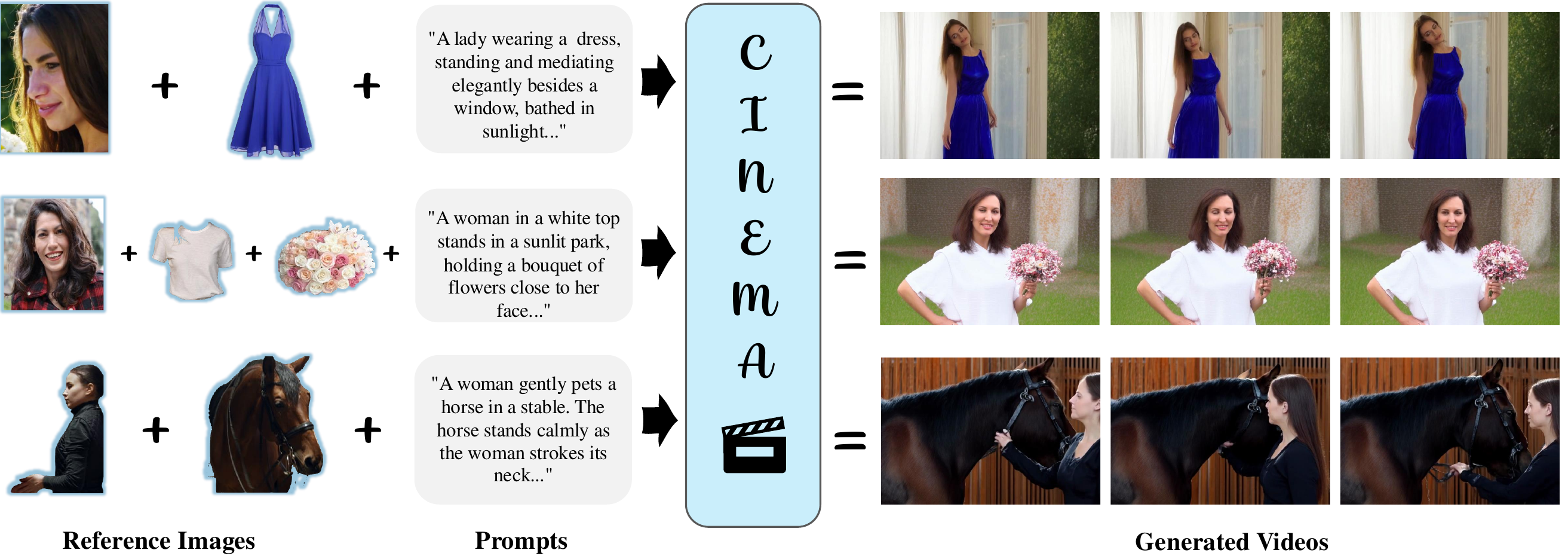}
    \captionof{figure}{We introduce \sysname, a video generation framework conditioned on a set of reference images and text prompts. \sysname enables the generation of videos visually consistent across multiple subjects from diverse scenes, providing enhanced flexibility and precise control over the video synthesis process.
 }
    \label{fig:teaser}
\end{center}%
}]

\maketitle
\begin{abstract}
Video generation has witnessed remarkable progress with the advent of deep generative models, particularly diffusion models. While existing methods excel in generating high-quality videos from text prompts or single images, personalized multi-subject video generation remains a largely unexplored challenge. This task involves synthesizing videos that incorporate multiple distinct subjects, each defined by separate reference images, while ensuring temporal and spatial consistency. Current approaches primarily rely on mapping subject images to keywords in text prompts, which introduces ambiguity and limits their ability to model subject relationships effectively. In this paper, we propose \sysname, a novel framework for coherent multi-subject video generation by leveraging Multimodal Large Language Model (MLLM). Our approach eliminates the need for explicit correspondences between subject images and text entities, mitigating ambiguity and reducing annotation effort. By leveraging MLLM to interpret subject relationships, our method facilitates scalability, enabling the use of large and diverse datasets for training. Furthermore, our framework can be conditioned on varying numbers of subjects, offering greater flexibility in personalized content creation. Through extensive evaluations, we demonstrate that our approach significantly improves subject consistency, and overall video coherence, paving the way for advanced applications in storytelling, interactive media, and personalized video generation.
\end{abstract}

\section{Introduction}
\label{sec:intro}

Advances in deep generative models, particularly diffusion models~\cite{ho2020denoising,song2020score,peebles2023scalable}, have significantly enhanced the ability to generate realistic and coherent videos from text prompts or images, attracting considerable attention from both academia and industry. Pioneering models such as Stable Video Diffusion~\cite{blattmann2023stable}, Runway Gen~\cite{runway2024runway}, Pika~\cite{pika2024sceneingredients}, and Sora~\cite{openai2024sora} have demonstrated remarkable capabilities in generating high-quality videos, continually expanding the boundaries of visual content creation. These models leverage large-scale datasets, advanced diffusion techniques, and scaling model size to produce visually compelling and temporally consistent results, establishing video generation as a cornerstone of modern generative AI.

Typically, video generation is driven by a user provided prompt (Text-to-Video), or a single image as the starting frame (Image-to-Video). 
Furthermore, personalized video generation aims to preserve the visual consistency of a given subject while seamlessly following the prompt's instructions.
Single-subject video generation has been extensively studied. For example, recent works like CustomCrafter~\cite{wu2024customcrafter} adapts to a given subject using LoRA~\cite{hu2022lora}, while DreamVideo-2~\cite{wei2024dreamvideo} and SUGAR~\cite{zhou2024sugar} enhance identity preservation through masking mechanism in the input image or attention operations, respectively.

In contrast, personalized multi-subject video generation remains a significantly challenging and underexplored task. It requires generating a video that features multiple subjects, each defined by distinct reference images, while optionally being guided by a text prompt to control scene context or dynamics. While existing models excel at generating videos from a single input modality, they often struggle to seamlessly integrate multiple subjects in a coherent and visually consistent manner. This limitation underscores a key challenge: effectively understanding the relationships between multiple subjects, their interactions, and their integration into a unified scene while preserving both temporal and spatial consistency.

Recent works~\cite{wang2024customvideo,liu2025phantom,chen2025multi,huang2025conceptmaster} have sought to address this challenge. 
VideoAlchemy~\cite{chen2025multi} encodes multiple subject images into tokens leveraging a frozen image encoder, binds them with corresponding word tokens from the text prompt, and incorporates these combined embeddings into the Diffusion Transformer~\cite{peebles2023scalable} via separate cross-attention layers.
Grounding visual concepts to entity words, however, is challenging~\cite{rohrbach2016grounding} and yet not practical in the era of scaling law. For example, in a video with “two persons are talking,” both visual subjects map to “two persons,” causing ambiguity. Furthermore, these methods struggle to establish meaningful correlations between the visual features of subjects under the context of prompt, resulting in a limited understanding of subject relationships. For example, in~\cite{chen2025multi}, visual tokens are merely concatenated after being linked to word tokens, without deeper semantic understanding or comprehension, leading to disorganized spatial arrangements and ambiguous subject interactions for generation.



In this paper, we introduce a novel framework for multi-subject video generation that leverages Multimodal Large Language Model (MLLM) guidance to ensure coherence and subject consistency.
We propose to exploit the comprehension capability of MLLM to interpret and orchestrate the relationships between multiple subjects, enabling the generation of personalized, multi-subject videos that are both visually coherent and contextually meaningful. 
Our approach is model-agnostic, and we seamlessly build \sysname upon a pre-trained, open-source video generative model~\cite{yang2024cogvideox}, highlighting its versatility and adaptability. Since most existing open-sourced video generative models~\cite{wan2.1,yang2024cogvideox,opensora,allegro2024} relies on LLM like~\cite{raffel2020t5}, and are not inherently compatible with MLLM, we introduce AlignerNet, a module designed to bridge this gap by aligning MLLM outputs with the native text features. Additionally, to ensure the preservation of each subject's visual appearance, we additionally inject Variational Autoencoder (VAE) features of reference images into our training, which serve as auxiliary conditioning signal and proves to be critical for maintaining consistency across frames.

\noindent To summarize our contribution:
\begin{itemize}
\item We propose \sysname, the first video generative model leveraging MLLM for multi-subject video generation task.
\item We demonstrate that our proposed method eliminates the need for training with explicit correspondences between subject images and entity words, enables easy scalability, allowing the use of large, diverse datasets for training.
\item With the comprehension capability of MLLM, we show that \sysname significantly improves multi-subject consistency, and overall video coherence in generation.
\end{itemize}

\section{Related Work}
\label{sec:related_work}

\paragraph{Video foundation model.}
\label{subsec:videogen}
Recent advancements in video generation often rely on Variational Autoencoders (VAEs)~\cite{wfvae, kingma2013auto, van2017neural} to compress raw video data into a low-dimensional latent space. Within this compressed latent space, large-scale generative pre-training is conducted using either diffusion-based methods~\cite{ho2020ddpm, song2021scorebased} or auto-regressive approaches~\cite{yu2023magvit, pmlr-v119-chen20s, ren2025videoworld}. Leveraging the scalability of Transformer models~\cite{vaswani2017attention, peebles2023scalable}, these methods have demonstrated steady performance improvements~\cite{videoworldsimulators2024, yang2024cogvideox, blattmann2023svd, guo2024i2v}.

Early methods for text-to-video (T2V) generation such as MagicTime~\cite{magictime, chronomagic} and AnimateDiff~\cite{guo2023animatediff} have laid the groundwork by integrating temporal modules into the 2D U-Net architecture~\cite{ronneberger2015unet}. These approaches typically involve redesigning and extensively training U-Net models to learn temporal dynamics, addressing the challenges posed by the dynamic nature of video content. Recent works, such as CogVideoX~\cite{yang2024cogvideox} and HunyuanVideo~\cite{kong2024hunyuanvideo}, have adopted more advanced architecture designs. First, they leverage a 3D causal VAE~\cite{yu2023language} for encoding and decoding raw video data, significantly improving the compression ratio of the original data. Second, they utilize the DiT~\cite{peebles2023scalable} model architecture, known for its superior scaling capabilities, and incorporate a 3D full-attention mechanism. This enhances the model's ability to capture temporal dynamics in video data, leading to substantial progress in T2V generation.

The image-to-video (I2V) generation task focuses on generating videos conditioned on a static input image. One widely adopted strategy involves integrating CLIP~\cite{radford2021clip} embeddings of the input image into the diffusion process. For example, VideoCrafter1~\cite{chen2023videocrafter1} and I2V-Adapter~\cite{guo2024i2v} leverage a dual cross-attention layer, inspired by the IP-Adapter~\cite{ye2023ip}, to effectively fuse these embeddings. Another prominent strategy involves expanding the input channels of diffusion models to concatenate the static image with noisy video frames. Methods such as SEINE~\cite{chen2023seine} and PixelDance~\cite{zeng2024make} achieve superior results by integrating image information more effectively into the video generation process. Additionally, hybrid approaches combine channel concatenation with attention mechanisms to inject image features. Notable examples include DynamiCrafter~\cite{xing2024dynamicrafter}, I2VGen-XL~\cite{zhang2023i2vgen}, and Stable Video Diffusion (SVD)~\cite{blattmann2023svd}. These methods aim to maintain consistency in both global semantics and fine-grained details. By utilizing this dual approach, they ensure that the generated videos remain faithful to the input image while exhibiting realistic and coherent motion dynamics.

\paragraph{Subject-driven image and video generation.}
\label{subsec:subjectgen}
Generating identity-consistent images and videos based on given reference images requires the model to accurately capture the ID features. Previous methods can be broadly divided into two categories: tuning-based and tuning-free solutions.

Tuning-based methods~\cite{zhou2024sugar, wei2024dreamvideo, chen2025multi, wu2024customcrafter} typically utilize efficient fine-tuning techniques, such as LoRA~\cite{hu2022lora} or DreamBooth~\cite{ruiz2023dreambooth}, to adapt pre-trained models. These approaches embed identity-specific information while retaining the original generative capabilities of the foundation model. However, these methods require fine-tuning for each new identity, which limits their real-world practicality.

In contrast, tuning-free methods adopt a feed-forward inference manner, eliminating the need for fine-tuning with new identity input. StoryDiffusion~\cite{zhou2025storydiffusion} employs a consistent self-attention mechanism and a semantic motion predictor to ensure smooth transitions and consistent identity. MS-Diffusion~\cite{wang2024ms} introduces a grounded resampler and a multi-subject cross-attention mechanism to extract detailed features of the reference subjects. ConsisID~\cite{yuan2024identity}, maintains facial identity consistency in videos through frequency decomposition. Specifically, it decomposes identity features into low-frequency and high-frequency signals, which are then injected into specific layers of the DiT~\cite{peebles2023scalable} model. Finally, ConceptMaster~\cite{huang2025conceptmaster} employs a CLIP~\cite{radford2021learning} image encoder and a Q-Former~\cite{li2023blip} to extract visual embeddings. It further fuses the visual and textual embeddings using a Decoupled Attention Module (DAM) and incorporates these embeddings into the diffusion model via a multi-concept injector, enabling multi-concept video customization.


\label{sec:method}

\begin{figure*}
  \centering
  \includegraphics[width=\linewidth]{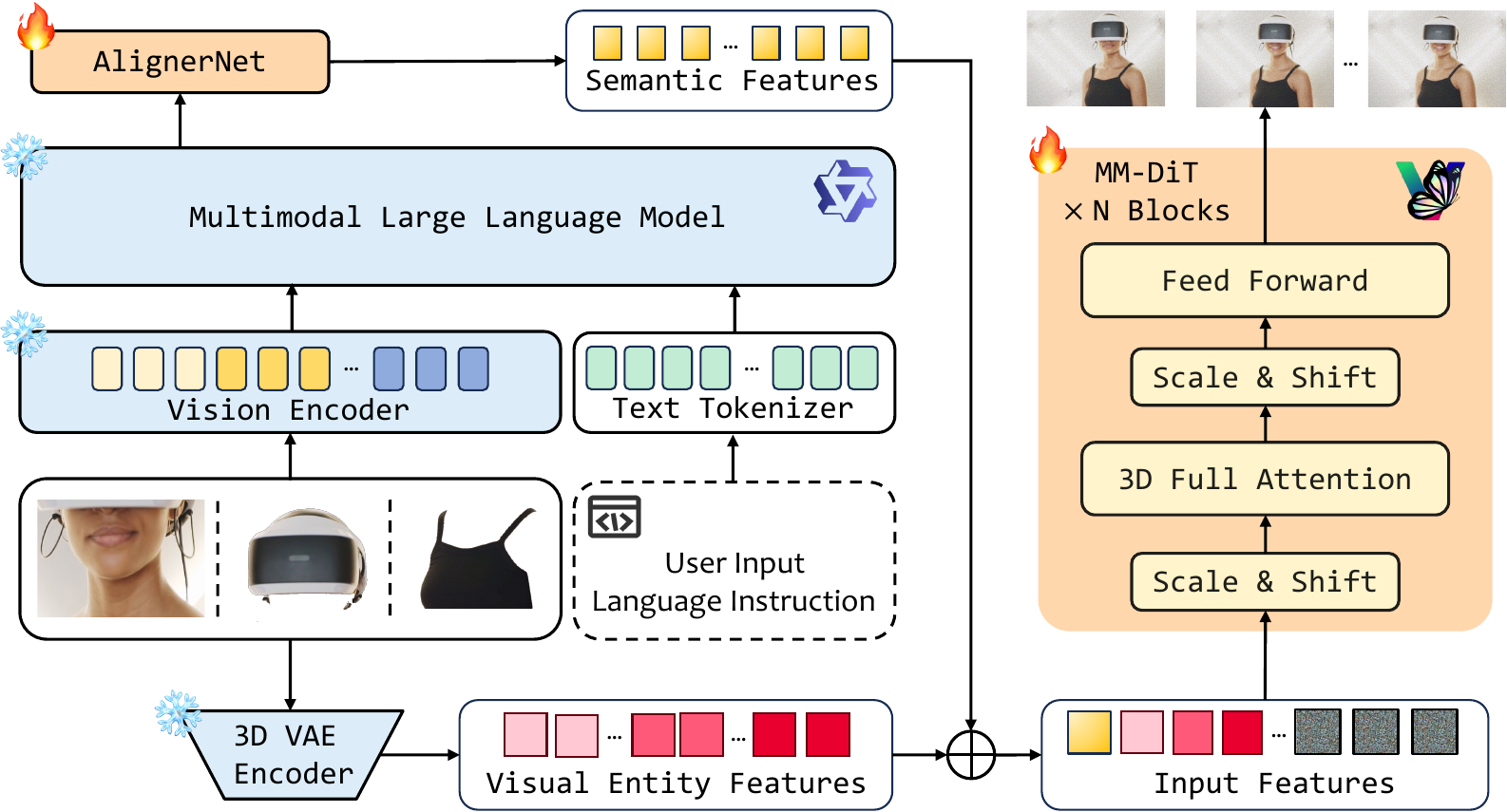}
  \caption{Our overall pipeline consists of a Multimodal Large Language Model, a semantic alignment network (AlignerNet), a visual entity encoding network, and a Diffusion Transformer that integrates the embeddings encoded by these two modules. $\oplus$ denotes concatenation.}
  \label{fig:overview}
\end{figure*}

\section{Method}

Given a set of reference images and corresponding text prompts, our objective is to generate a video that maintains the consistent appearance of the subject across frames. This task presents several challenges, as it requires preserving the identity depicted in the reference images, faithfully adhering to the instructions provided in the text prompts, and ensuring temporal coherence throughout the entire video sequence.

\subsection{Preliminaries}
\label{subsec:preliminaries}

\paragraph{Multimodal large language models.}
Multimodal large language models (MLLMs)~\cite{bai2025qwen2, li2023blip, team2023gemini, liu2023visual, wang2024qwen2} extend traditional large language models by enabling cross-modal understanding and generation across various modalities, such as text, image, video, and audio. Building on the success of unimodal LLMs like GPT~\cite{brown2020language} and PaLM~\cite{chowdhery2023palm}, MLLMs incorporate modality-specific encoders, such as CLIP~\cite{radford2021learning} for visual input and Whisper~\cite{radford2023robust} for audio, to project heterogeneous inputs into a unified embedding space. These embeddings are well-aligned with cross-modal fusion modules, such as attention-based mechanisms like Perceiver Resamplers~\cite{jaegle2021perceiver}, and processed by a decoder-only transformer backbone adapted for multimodal tasks via parameter-efficient fine-tuning~\cite{hu2022lora, ruiz2023dreambooth}.

\paragraph{Multimodal diffusion transformer.}
The Multimodal Diffusion Transformer (MM-DiT) is an advanced transformer-based diffusion backbone introduced by ~\cite{esser2024scaling}, designed to enhance multimodal alignment and generation consistency in generative tasks. Specifically, MM-DiT employs a multimodal joint attention mechanism that facilitates efficient visual-textual cross-modal interaction. This approach has been shown to significantly outperform cross-attention-based models in terms of generation quality, text alignment, and computational efficiency.




\subsection{Data Curation}
We present a data processing pipeline that efficiently ingests training videos, generates text labels, and extracts references—such as humans, faces, and objects—facilitating comprehensive video-based tasks.

To ensure a clean, and high-quality training datasets, we perform rigorous data pre-processing and filtering. Given a set of training videos, we first perform scene-change detection to segment each video into multiple clips, \ie, $ \videoclip_{1}, \videoclip_{2}, \ldots, \videoclip_{n}$. Following previous video generation approaches~\cite{yang2024cogvideox, opensora}, we evaluate the aesthetics and motion magnitude of each clip $\videoclip_{i}$, discarding those with poor aesthetic quality or minimal motion.

For each filtered video clip $\videoclip_{i}$, we exploit Qwen2-VL~\cite{wang2024qwen2} to generate a caption that depicts its overall content $\captions$ focusing on motion aspect. We then prompt the MLLM to identify and list the objects (entity words) mentioned in the caption as clip-level labels, such as \emph{``cat''}, \emph{``hat''}, and \emph{``t-shirt''}.
For each object label, we compute a bounding box in the first frame of the clip using GroundingDINO~\cite{liu2024grounding}, then continuously segment the object with SAM2~\cite{ravi2024sam2}.These segmented subject maps serve as reference image $\referenceimage^{\text{Obj}}_{i,k}$.

Maintaining consistent appearance of humans is notoriously challenging in image/video generation task. To ensure a clear representation of humans, we first detect humans and faces in the initial frame using YOLO~\cite{redmon2016yolo}, segment the human regions with SAM2, and apply a slightly looser bounding box around faces to avoid including irrelevant elements. These segmented images are then stored as $\referenceimage^{\text{Human}}{i}$ and $\referenceimage^{\text{Face}}{i}$, respectively.

Consequently, each training input comprises a set of object segmentations, human segmentations, and cropped faces, together with their text labels. We define the training data as 
$
\mathcal{R}_i = \{\,\captions_i,\, \referenceimage^{\text{Human}}_{i},\, \referenceimage^{\text{Face}}_{i},\, \referenceimage^{\text{Obj}}_{i,1},\, \referenceimage^{\text{Obj}}_{i,2},\, \ldots,\, \referenceimage^{\text{Obj}}_{i,k} \}.
$
For each training example, we pair this set with the corresponding video clip $\videoclip_i$.

\subsection{Video Generation via MLLM-Based Guidance}
\label{subsec:model_design}
We propose a novel framework, \sysname, for coherent multi-subject video generation, leveraging the powerful comprehending capabilities of MLLMs.
Utilizing MM-DiT for video generation, \sysname integrates multimodal conditional information through three key modules, \ie, \textit{multimodal large language model}, \textit{semantic alignment network}, and \textit{visual entity encoding}.
An overview of our framework is provided in Figure~\ref{fig:overview}.
The MLLM is employed to encode multimodal conditions (\eg, images and text) into unified feature representations. Additionally, the semantic alignment network AlignerNet is introduced to bridge the gap between this unified representation space and the original DiT conditioning feature space. Lastly, the visual entity encoding is designed to capture fine-grained entity-level visual attributes. The architecture and functionality of each module are elaborated in the following subsection.


\begin{figure}
    \centering
    \includegraphics[width=\linewidth]{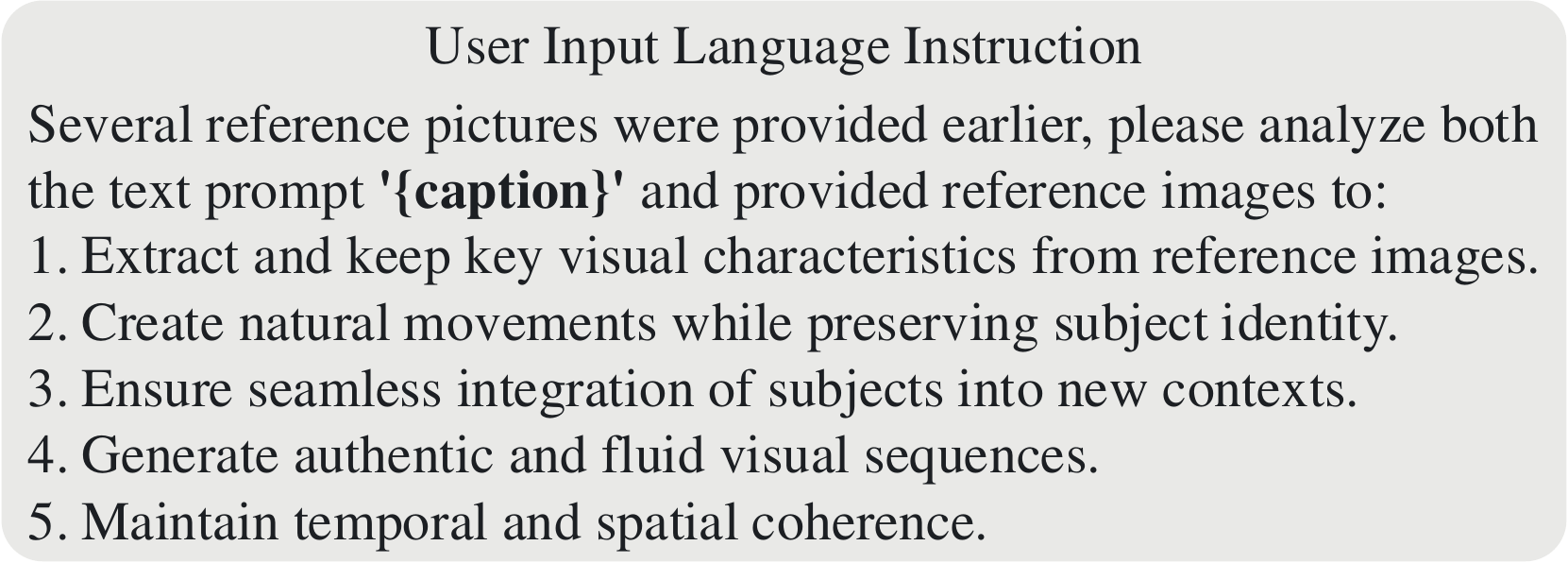} 
    \caption{Our language instruction template for MLLM.}
    \label{fig:instruction}
\end{figure}

\paragraph{Multimodal large language model.}
At the core of \sysname lies the utilization of an MLLM.
Specifically, Qwen2-VL~\cite{wang2024qwen2} is adopted to generate unified representations from arbitrary reference images, 
and text prompts.
These unified representations are then fed into the MM-DiT backbone, where multimodal joint attention mechanism facilitates interactive fusion of the encoded tokens. In addition, we employ an instruction tuning approach, incorporating a specialized instruction template to guide the MLLM for more accurate and contextually relevant encoding results. The specific instruction is as shown in Figure~\ref{fig:instruction}.

\paragraph{Semantic alignment network.}
We build \sysname on the basis of pre-trained, open-sourced video generative model model, CogVideoX~\cite{yang2024cogvideox}, which is largely pre-trained on text/image-to-video task, encoding text prompts with a large language model~\cite{raffel2020t5}. To address the semantic representation gap across LLM and MLLM, we propose to replace T5 encoder with MLLM by introducing AlignerNet module.

Our AlignerNet is a transformer-based network. In particular, AlignerNet maps the hidden states generated by the MLLM onto the feature space of the T5 text encoder, ensuring that the resulting multimodal semantic features encompassing both visual and textual information are well-aligned with the feature space of the T5 encoder. We use the combination of Mean Squared Error (MSE) and Cosine Similarity loss to optimize the AlignerNet.
Specifically, let $\tfivefeatures \in \mathbb{R}^{K\times d}$ represents the feature vectors generated by T5 encoder, and $\mllmfeatures \in \mathbb{R}^{K\times d}$ are the hidden states of MLLM after mapping by AlignerNet. The MSE loss serves to minimize the Euclidean distance between $\tfivefeatures$ and $\mllmfeatures$ as follows:
\begin{equation}
    \label{eq:loss_mse}
    \lossmse = \frac{1}{N} \sum_{i=1}^{N} \| \mllmfeatures^{(i)} - \tfivefeatures^{(i)} \|_2^2.
\end{equation}

The Cosine Similarity loss focuses on aligning the directions of $\tfivefeatures$ and $\mllmfeatures$. It is given by:
\begin{equation}
    \label{eq:loss_cosine}
    \losscos = \frac{1}{N} \sum_{i=1}^{N} \left( 1 - \frac{\langle \mllmfeatures^{(i)}, \tfivefeatures^{(i)} \rangle}{\| \mllmfeatures^{(i)} \|_2 \cdot \| \tfivefeatures^{(i)} \|_2} \right).
\end{equation}

Finally, the overall loss function is shown in Eq.~\ref{eq:loss_all}, where $\lambda_{\mathrm{mse}}$ and $\lambda_{\mathrm{cos}}$ are coefficients to adjust the weights between the two loss terms:
\begin{equation}
    \label{eq:loss_all}
    \loss_{\text{AlignerNet}} = \lambda_{\mathrm{mse}} \cdot \lossmse + \lambda_{\mathrm{cos}} \cdot \losscos.
\end{equation}

\paragraph{Visual entity encoding.}
We observe that the visual features encoded by the pre-trained MLLM~\cite{wang2024qwen2} are coarse-grained, as its training objective is to minimize a similarity loss between visual and text representations, primarily captures semantic-level image features. However, our task requires fine-grained visual identity (ID) features to handle arbitrary reference images as conditions. To bridge this gap, we use a VAE encoder that performs entity-level encoding on each reference image to enhance visual ID consistency.
Specifically, as shown in Figure~\ref{fig:overview}, given a variable number of reference images, the VAE first encodes each image to produce latent features. These features are then flattened into a sequence of embeddings, padded to a fixed length (denoted as $\visualfeatures \in \mathbb{R}^{M \times d}$), and fed into MM-DiT. 

\paragraph{Multimodal feature conditioned denoising.}
After obtaining the multi-modality features  $\mllmfeatures$ and visual features $\visualfeatures$, we concatenate these tokens to form a unified multimodal feature $\unifiedfeatures$ as follows:
\begin{equation}
    \unifiedfeatures = \text{Concat}(\mllmfeatures, \visualfeatures) \in \mathbb{R}^{(K + M) \times d}.
\end{equation}
Next, the unified multimodal features $\unifiedfeatures$ and the noisy tokens $\noisetokens$ on timestep $t$ are concatenated to form the final input sequence of MM-DiT backbone:
\begin{equation}
    \inputfeatures = \text{Concat}(\unifiedfeatures, \noisetokens) \in \mathbb{R}^{(K + M + T) \times d}.
\end{equation}

\begin{figure}
    \centering
    \includegraphics[width=\linewidth]{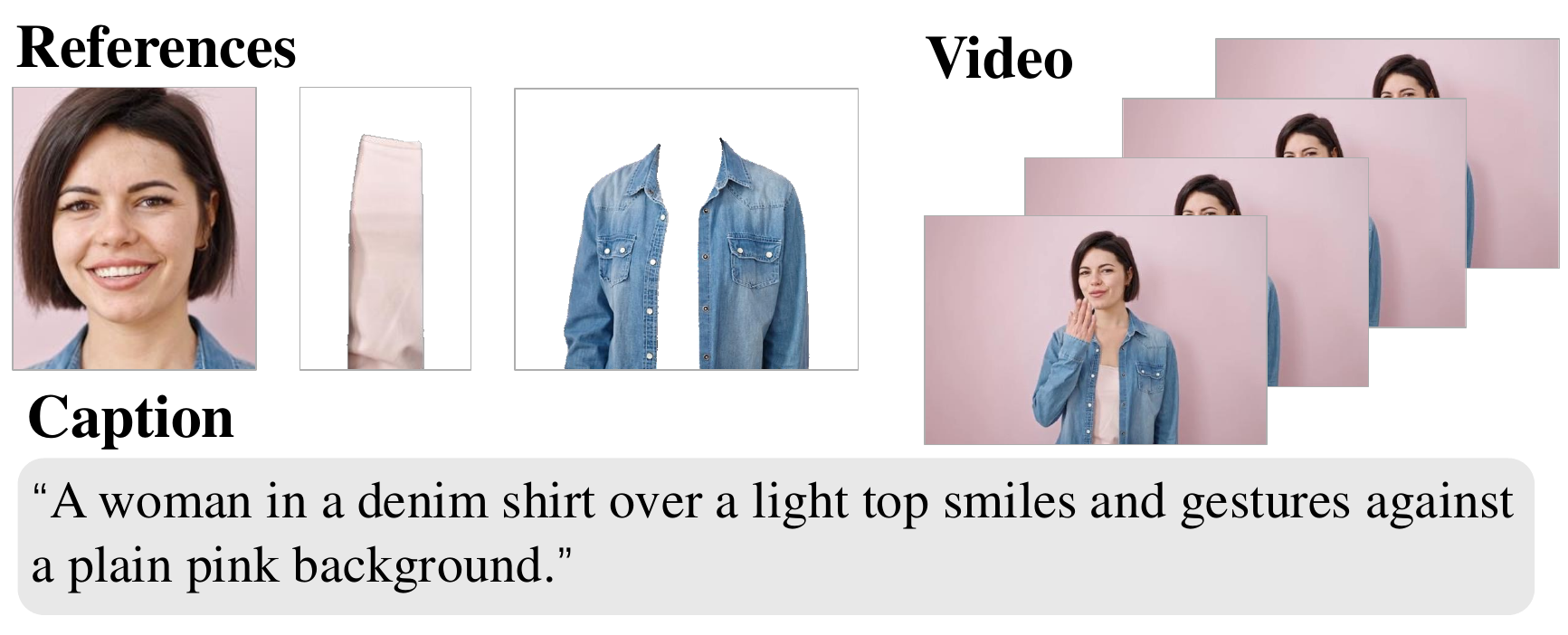} 
    \caption{An example of reference images, along with the corresponding video and caption, from our training set.}
    \label{fig:training_example}
\end{figure}

A dual-branch transformer model~\cite{yang2024cogvideox} is then employed, taking $\unifiedfeatures$ into the text branch and $ \noisetokens$ to the vision branch. It features a joint attention mechanism that dynamically fuses multimodal-enhanced feature tokens as follows:
\begin{equation}
    \text{Attention}(\mathbf{Q},\mathbf{K},\mathbf{V})=\text{softmax}\left(\frac{\mathbf{Q}^i (\mathbf{K})^T}{\sqrt{d}}\right)\mathbf{V},
\end{equation}
where $\mathbf{Q}=\inputfeatures \mathbf{W}_Q$, $\mathbf{K}=\inputfeatures \mathbf{W}_K$, and $\mathbf{V}=\mathbf{\inputfeatures} \mathbf{W}_V$. Both $\mathbf{W}_Q$, $\mathbf{W}_K$, and $\mathbf{W}_V$ are trainable weight matrices to encode the input feature maps. During the attention operation, the sequences from both modalities are concatenated, and joint attention weights are computed to achieve complementary information exchange.

Finally, the MM-DiT is fully finetuned on the diffusion~\cite{ho2020ddpm} training objective as:
\begin{equation}
    \loss = \mathbf{E}_{t, x_0, \epsilon} || \epsilon - \epsilon_\theta (\sqrt{\baralphat} x_0 + \sqrt{1-\baralphat} \epsilon, t) ||^2 ,
\end{equation}
where $t$ is sampled by explicit uniform sampling.

\section{Experiments}

\subsection{Experimental Setup}

\begin{figure*}
  \centering
  \includegraphics[width=\linewidth]{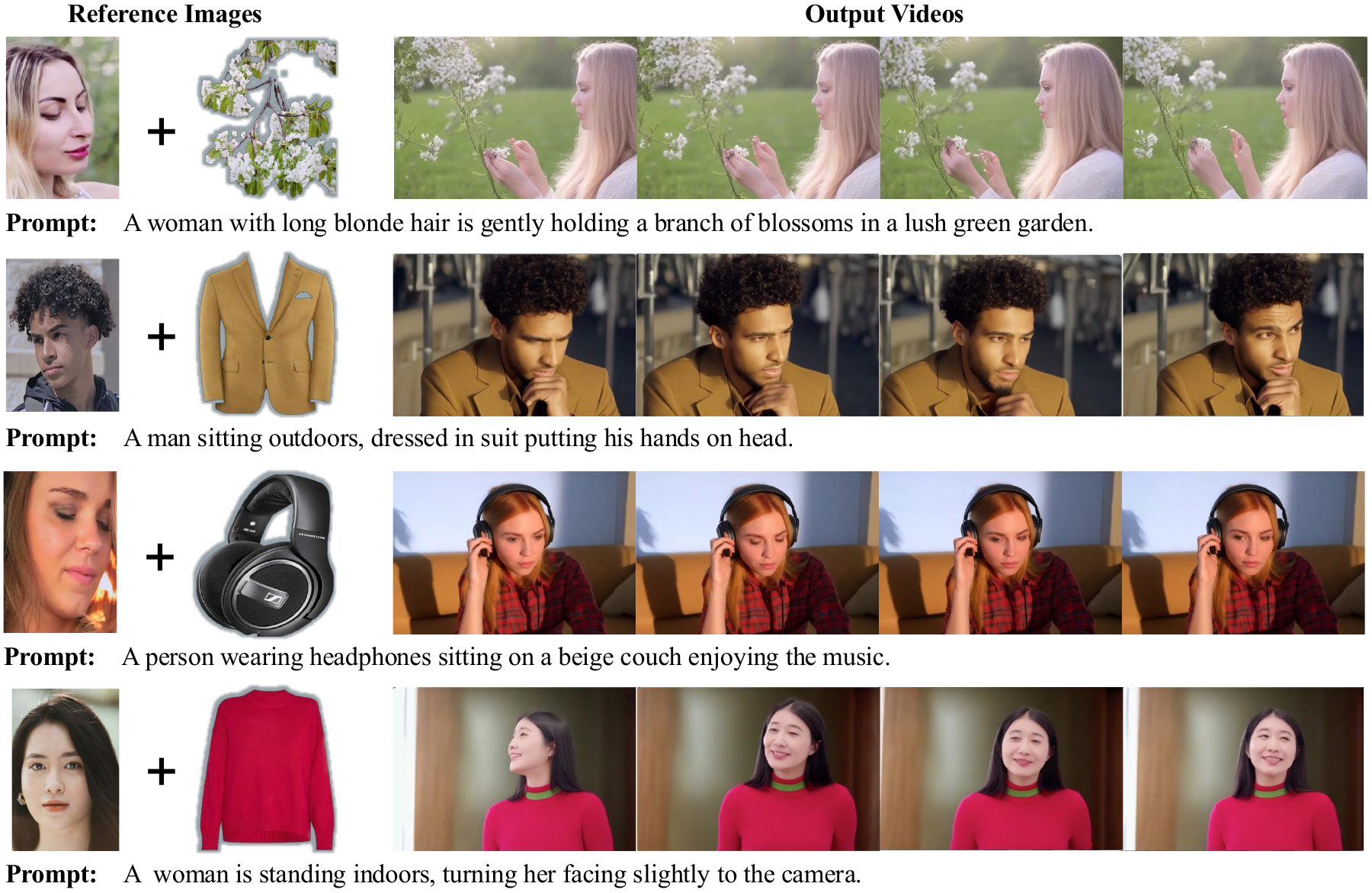}
  \caption{Qualitative evaluation of our method. The reference images are shown on the left, along with the text prompt at the bottom. In each case, we show four frames uniformly sampled from the generated 45-frame video.}
  \label{fig:qualitative_results}
\end{figure*}

\paragraph{Dataset.}
We train our model on a self-curated, high-quality video dataset 
(see the examples in Figure~\ref{fig:training_example}).
Specifically, the original training set contains approximately 5.1 million videos. After filtering, we obtain about 1.46 million video clips, each paired with one to six human and object references. Each video clip consists of 96 frames, and
all the paired references are stored as RGBA images.

\paragraph{Network components.}
For the video diffusion model, we use an open-source version of the image-to-video pretrained DiT model based on CogVideoX~\cite{yang2024cogvideox} as the base model. Due to insufficient training, the original I2V model does not converge effectively. Therefore, we continuously train the~\cite{yang2024cogvideox} model on I2V task first. For MLLM, we adopt Qwen2-VL-7B-Instruct~\cite{wang2024qwen2}, and freeze its parameters during the training of the entire framework. We adopt an attention-based architecture for AlignerNet, which consists of six attention layers with a hidden width of 768, eight attention heads, and 226 latent tokens to align with the original T5 sequence length. The input dimension is 2048, and the output dimension is aligned with the DiT.

\paragraph{Training strategies.}
We train our model using the AdamW optimizer with $\beta_1 = 0.9$, $\beta_2 = 0.95$, and a weight decay of $0.001$. 
The initial learning rate is set to $1 \times 10^{-5}$ and follows a cosine annealing schedule with restarts.
To stabilize early training, we employ a linear warm-up phase in the first 100 steps. The batch size is set to one. 
We apply gradient clipping with a maximum norm of 1.0 to prevent exploding gradients.
The conditional reference is randomly dropped with a probability of 0.05. 
Our model is trained with 128 NVIDIA H100 GPUs for two days. The input videos are processed at a resolution of 448P, with a sequence length of 45 frames and a temporal stride of two.
The AlignerNet is pre-trained with the MSE loss and cosine similarity loss to
match the feature distribution of the T5 model, with both $\lambda_{\mathrm{mse}}$ and $\lambda_{\mathrm{cos}}$ set to be $1$ in Eq.~\ref{eq:loss_all}.
Then we jointly train AlignerNet and DiT.

\begin{figure*}
  \centering
  \includegraphics[width=\linewidth]{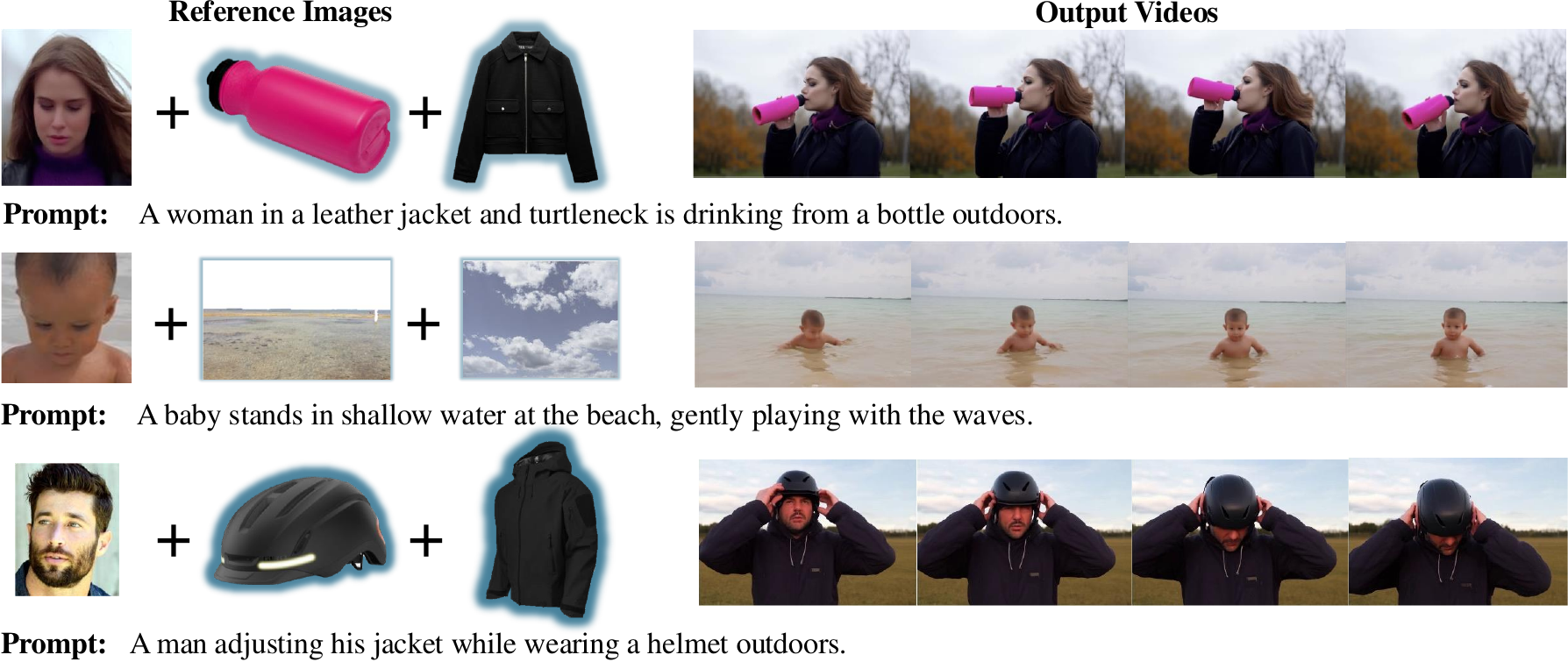}
  \caption{Qualitative evaluation of our method dealing with multiple concepts. Our model is capable of encoding and understanding multiple subjects based on the reference images.}
  \label{fig:qualitative_results_multi_concepts}
\end{figure*}


\subsection{Qualitative Results}

Figures \ref{fig:qualitative_results} and \ref{fig:qualitative_results_multi_concepts} demonstrate qualitative examples of our method, which synthesizes coherent videos from reference images paired with succinct textual descriptions. The results clearly indicate that our model successfully preserves the distinct visual attributes of the provided subjects, such as clothing textures, accessories, and environmental contexts. For instance, in Figure \ref{fig:qualitative_results}, the generated video accurately retains the blonde hair and gentle interaction with blossoms, precisely following the prompt. Similarly, the detailed texture and color of the brown suit worn by the man are faithfully reproduced along with natural gestures like putting his hands on his chin. Furthermore, Figure \ref{fig:qualitative_results_multi_concepts} showcases the capability of our model to handle complex scenarios with multiple visual concepts. An illustrative example includes the woman outdoors wearing a leather jacket and drinking from a distinctly colored bottle, where both the clothing and the object's visual details are accurately captured. Additionally, the baby's interaction with shallow beach water effectively demonstrates the model's skill in recreating natural movements consistent with the described environment. These examples confirm that our method can accurately represent interactions among individuals, objects, and settings, producing contextually relevant and visually convincing motions\footnote{The corresponding animations can be found in the composed video demo included in our supplementary materials.}.

\begin{figure*}
  \centering
  \includegraphics[width=\linewidth]{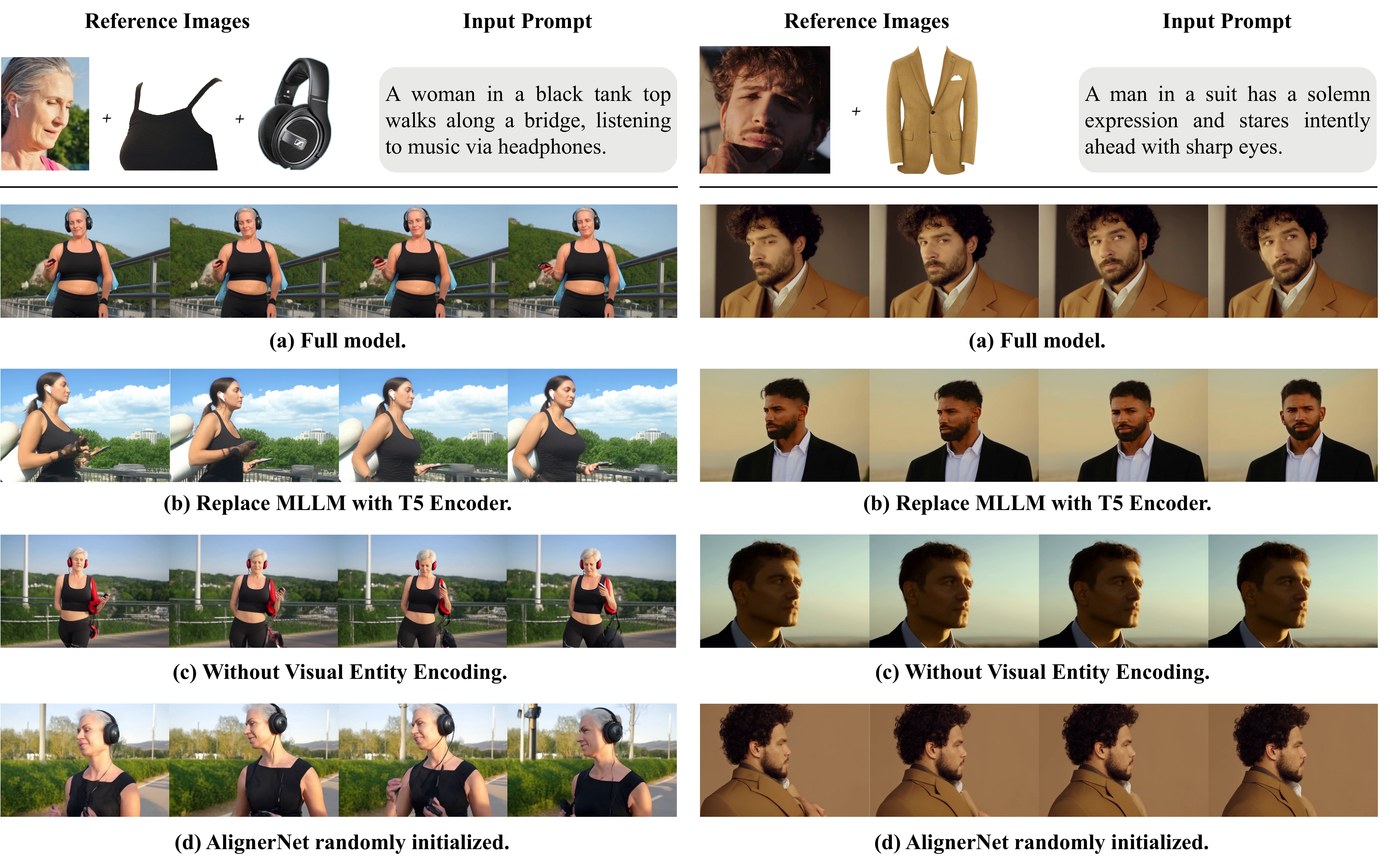}
  \caption{Qualitative comparison for ablation studies. The reference images and text prompt are shown on the top. The result of the full model is in the first line, followed by those from different ablation experiments. We show four frames uniformly sampled from the generated 45-frame video of each method.}
  \label{fig:ablation_study}
\end{figure*}


\subsection{Ablation Studies}

Several ablation studies are conducted to verify the effectiveness of each component in \sysname. The corresponding qualitative evaluation results are demonstrated in Figure~\ref{fig:ablation_study}.

\paragraph{Semantic alignment network.}
The proposed AlignerNet bridges the feature gap between the text representation from the MLLM~\cite{wang2024qwen2} and the T5 encoder~\cite{raffel2020t5}. To this end, AlignerNet is pretrained to minimize the MSE and cosine distance to the output of the T5 encoder. To demonstrate the necessity of pretraining, we conduct experiments by randomly initializing AlignerNet and training it end-to-end with MM-DiT. As shown in Figure~\ref{fig:ablation_study}(b), we observe unstable and inconsistent results and hypothesize that these issues might be due to the random initialization of AlignerNet, causing slower convergence, increased training difficulty, and poor alignment between text and reference image modalities. This finding underscores the critical importance of pretraining AlignerNet to effectively map the MLLM output to the T5's embedding space.

\paragraph{Visual entity encoding.}
This module generates image latents that preserve fine-grained visual details from the reference images. The MM-DiT then synthesizes videos with high subject consistency from reference images. In our ablation study, we replace the VAE-derived embeddings with features extracted from the vision encoder of the original MLLM~\cite{wang2024qwen2}. As demonstrated in Figure~\ref{fig:ablation_study}(c), the generated videos still contain the corresponding subjects; however, they exhibit variations in fine details or identity, particularly for human subjects. The generated video only keeps general attributes of the subjects. This result highlights the necessity of the VAE encoder in preserving visual details, ensuring high-level reference consistency in generated videos.

\paragraph{Unified multimodal encoder.}
This encoder serves as the core of our framework, integrating both text prompts and visual references into a cohesive representation. In our ablation study, we replace it with the standard T5 text encoder. Under this configuration, text prompts are encoded by T5, while reference images are processed separately using the VAE encoder, with no fusion between the two modalities. As shown in Figure~\ref{fig:ablation_study}(d), this results in videos with reduced consistency between visual and textual prompts. Specifically, the interactions described in the text prompt may not involve the intended subject from the reference images. This issue highlights the importance of unifying textual and visual inputs before feeding them into the MM-DiT for video generation.

\begin{figure}
    \centering
    \includegraphics[width=\linewidth]{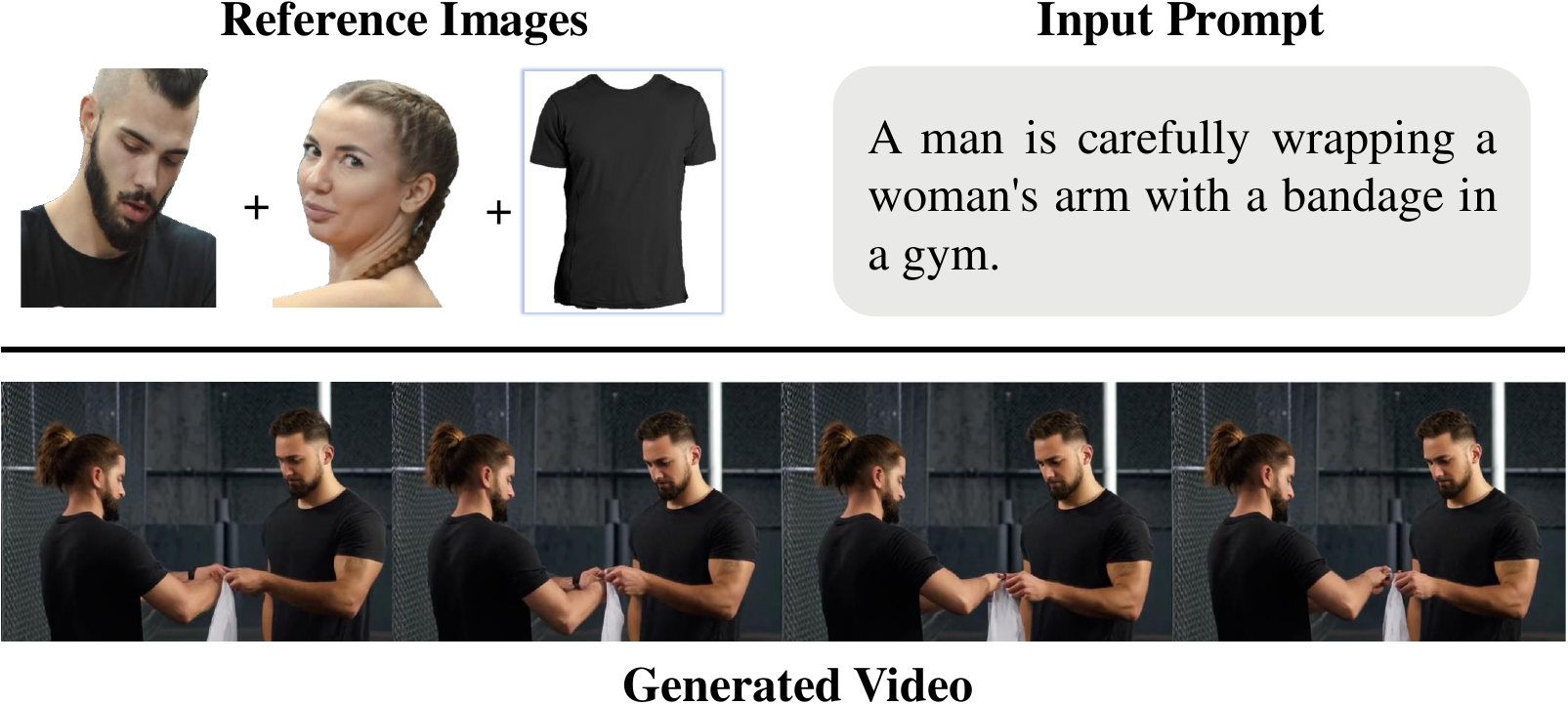} 
    \caption{A failure case of our method. The identities of the two persons are not clearly distinguishable or well-maintained in the generated video.}
    \label{fig:failure_case}
\end{figure}

\paragraph{Limitations.}
While \sysname is model-agnostic, its performance is predominantly constrained by the capabilities of the underlying video foundation model, which can impact both output quality and temporal consistency obviously. Furthermore, our method struggles to accurately distinguish between different human identities (as illustrated in Figure~\ref{fig:failure_case}), which hinders its ability to handle multiple individuals effectively, a common challenge in multi-subject image and video generation systems~\cite{ruiz2023dreambooth,ding2024freecustom}.

\section{Conclusion}

In this work, we introduce \sysname, a multi-subject video generation framework that leverages MLLM to further enhance coherence and contextual alignment. By employing MLLM as the encoder, it eliminates explicit subject-text correspondences required for training, and so our approach enables flexible and harmonious video synthesis, yielding improved subject consistency and interaction modeling. 

As part of future work, we aim to leverage more advanced video foundation models~\cite{kong2024hunyuanvideo,wan2.1}, to enhance the modeling of subject relationships, support dynamic appearances, and improve the differentiation of multi-human identities. These advancements will contribute to achieving more robust and controllable video generation.

{
    \small
    \bibliographystyle{ieeenat_fullname}
    \bibliography{main}
}


\end{document}